\documentclass[conference, 10pt]{IEEEtran}
\IEEEoverridecommandlockouts
\usepackage{xspace} 
\usepackage[numbers,sort&compress]{natbib}
\usepackage{amsmath,amssymb,amsfonts}
\usepackage{algorithmic}
\usepackage{graphicx}
\usepackage{textcomp}
\usepackage{url}            
\usepackage[breaklinks=true]{hyperref}
\hypersetup{
     colorlinks = true,
     linkcolor  =blue, 
     urlcolor   =blue,
     citecolor  = black
}
\usepackage[dvipsnames]{xcolor}
\def\BibTeX{{\rm B\kern-.05em{\sc i\kern-.025em b}\kern-.08em
    T\kern-.1667em\lower.7ex\hbox{E}\kern-.125emX}}

\graphicspath{{./}{./figures/}{../figures/}}
\DeclareGraphicsExtensions{.pdf,.png,.jpg}
\usepackage{siunitx}
\usepackage[capitalize]{cleveref}

\DeclareRobustCommand{\IEEEauthorrefmarks}[1]{\smash{\textsuperscript{\footnotesize #1}}}

\newcommand{\ie}{\mbox{i.\hspace{1pt}e.}\xspace}
\newcommand{\eg}{\mbox{e.\hspace{1pt}g.}\xspace}

\newcommand{\etc}{\mbox{etc.}}

\usepackage{xcolor}
	\definecolor{EricColor}{RGB}{0,49,60}
	
\begin{document}

\title{MLExchange: A web-based platform enabling exchangeable machine learning workflows for scientific studies
}

\author{
\IEEEauthorblockN{
Zhuowen Zhao\IEEEauthorrefmarks{1}\IEEEauthorrefmark{1},
Tanny Chavez\IEEEauthorrefmarks{1}\IEEEauthorrefmark{1},
Elizabeth A. Holman\IEEEauthorrefmarks{1},
Guanhua Hao\IEEEauthorrefmarks{1}, 
Adam Green\IEEEauthorrefmarks{1}, 
Harinarayan Krishnan\IEEEauthorrefmarks{1, 2}, \\  
Dylan McReynolds\IEEEauthorrefmarks{1},
Ronald J. Pandolfi\IEEEauthorrefmarks{2},
Eric J. Roberts\IEEEauthorrefmarks{2, 3}, 
Petrus H. Zwart\IEEEauthorrefmarks{2, 3}, 
Howard Yanxon\IEEEauthorrefmarks{4}, \\
Nicholas Schwarz\IEEEauthorrefmarks{4}, 
Subramanian Sankaranarayanan\IEEEauthorrefmarks{5, 6},
Sergei V. Kalinin\IEEEauthorrefmarks{7}, 
Apurva Mehta\IEEEauthorrefmarks{8}, 
Stuart I. Campbell\IEEEauthorrefmarks{9} \\ and 
Alexander Hexemer\IEEEauthorrefmarks{1}\IEEEauthorrefmark{1}} \\
\IEEEauthorblockA{\IEEEauthorrefmarks{1}Advanced Light Source (ALS) Division,
Lawrence Berkeley National Laboratory, Berkeley, CA 94720}
\IEEEauthorblockA{\IEEEauthorrefmarks{2}Center for Advanced Mathematics for Energy Research Applications (CAMERA), \\
Lawrence Berkeley National Laboratory, Berkeley, CA 94720}
\IEEEauthorblockA{\IEEEauthorrefmarks{3}Molecular Biophysics and Integrated Bioimaging Division (MBIB), \\
Lawrence Berkeley National Laboratory, Berkeley, CA 94720}
\IEEEauthorblockA{\IEEEauthorrefmarks{4}Advanced Photon Source, 
Argonne National Laboratory, Lemont, IL 60439}
\IEEEauthorblockA{\IEEEauthorrefmarks{5}Center for Nanoscale Materials (CNM),
Argonne National Laboratory, Lemont, IL 60439}
\IEEEauthorblockA{\IEEEauthorrefmarks{6}Department of Mechanical and Industrial Engineering,
University of Illinois Chicago, Chicago, IL 60607}
\IEEEauthorblockA{\IEEEauthorrefmarks{7}Center for Nanophase Materials Sciences,
Oak Ridge National Laboratory, Oak Ridge, TN 37830} 
\IEEEauthorblockA{\IEEEauthorrefmarks{8}SLAC National Accelerator Laboratory, Menlo Park, CA 94025} 
\IEEEauthorblockA{\IEEEauthorrefmarks{9}National Synchrotron Light Source II, 
Brookhaven National Laboratory, Upton, NY 11973} 
\IEEEauthorblockA{\IEEEauthorrefmark{1}Email: zzhao2@lbl.gov, tanchavez@lbl.gov, ahexemer@lbl.gov}
}

\IEEEoverridecommandlockouts
\IEEEpubid{\begin{minipage}{\textwidth}\ \\[12pt]
  DOI: \href{https://ieeexplore.ieee.org/document/10024637}{\nolinkurl{10.1109/XLOOP56614.2022.00007}} \\
  © 20XX IEEE. Personal use of this material is permitted.  Permission from \\
  IEEE must be obtained for all other uses, in any current or future media, \\
  including reprinting/republishing this material for advertising or promotional\\
  purposes,creating new collective works, for resale or redistribution to servers \\
  or lists, or reuse of any copyrighted component of this work in other works.
\end{minipage}}
\maketitle
\IEEEpubidadjcol

\begin{abstract}
Machine learning (ML) algorithms are showing a growing trend in helping the scientific communities across different disciplines and institutions to address large and diverse data problems.
However, many available ML tools are programmatically demanding and computationally costly.
The MLExchange project aims to build a collaborative platform equipped with enabling tools that allow scientists and facility users who do not have a profound ML background to use ML and computational resources in scientific discovery. 
At the high level, we are targeting a full user experience where managing and exchanging ML algorithms, workflows, and data are readily available through web applications. 
Since each component is an independent container, the whole platform or its individual service(s) can be easily deployed at servers of different scales, ranging from a personal device (laptop, smart phone, \etc) to high performance clusters (HPC) accessed (simultaneously) by many users. 
Thus, MLExchange renders flexible using scenarios---users could either access the services and resources from a remote server or run the whole platform or its individual service(s) within their local network.

\end{abstract}


\section{Introduction}

\begin{figure*}
\centering
\includegraphics[width=\linewidth]{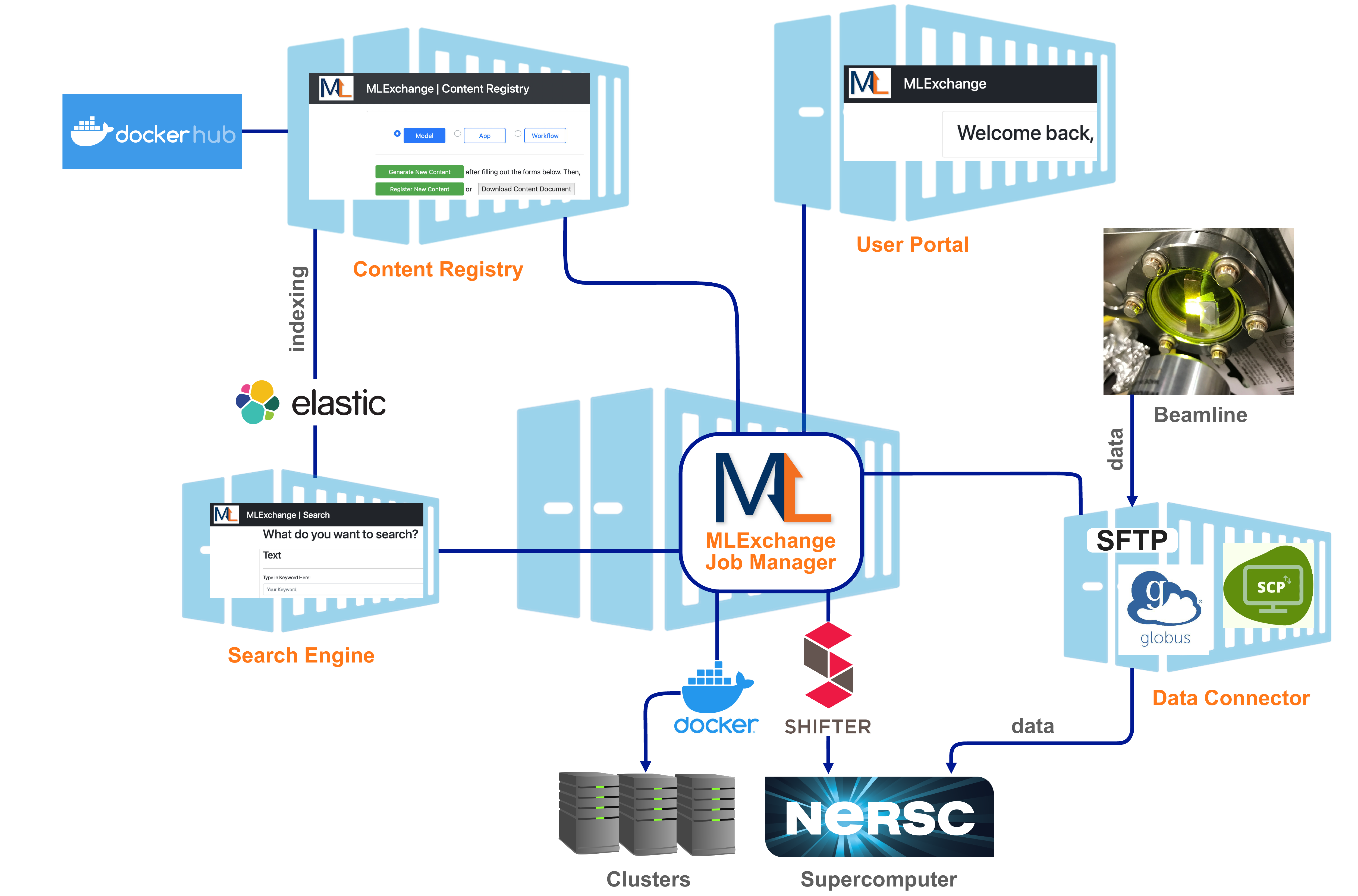}
\caption{Achieving scientific MLOps in MLExchange: user can (1) log in to the user portal and get authenticated; (2) use the data connector to securely transfer data to the designated location(s); (3) search for the relevant scientific models and applications; (4) launch the selected applications from the content registry and access the individual (interactive) application interface through the corresponding URL. All MLExchange components are containerized software. The job manager is a \emph{central} coordinator that handles job requests from the other components. A job manager should be deployed on a machine to leverage the machine's computing resources (CPU and GPU).}
\label{fig: architecture}
\end{figure*}

\begin{figure*}
\centering
\includegraphics[width=\linewidth]{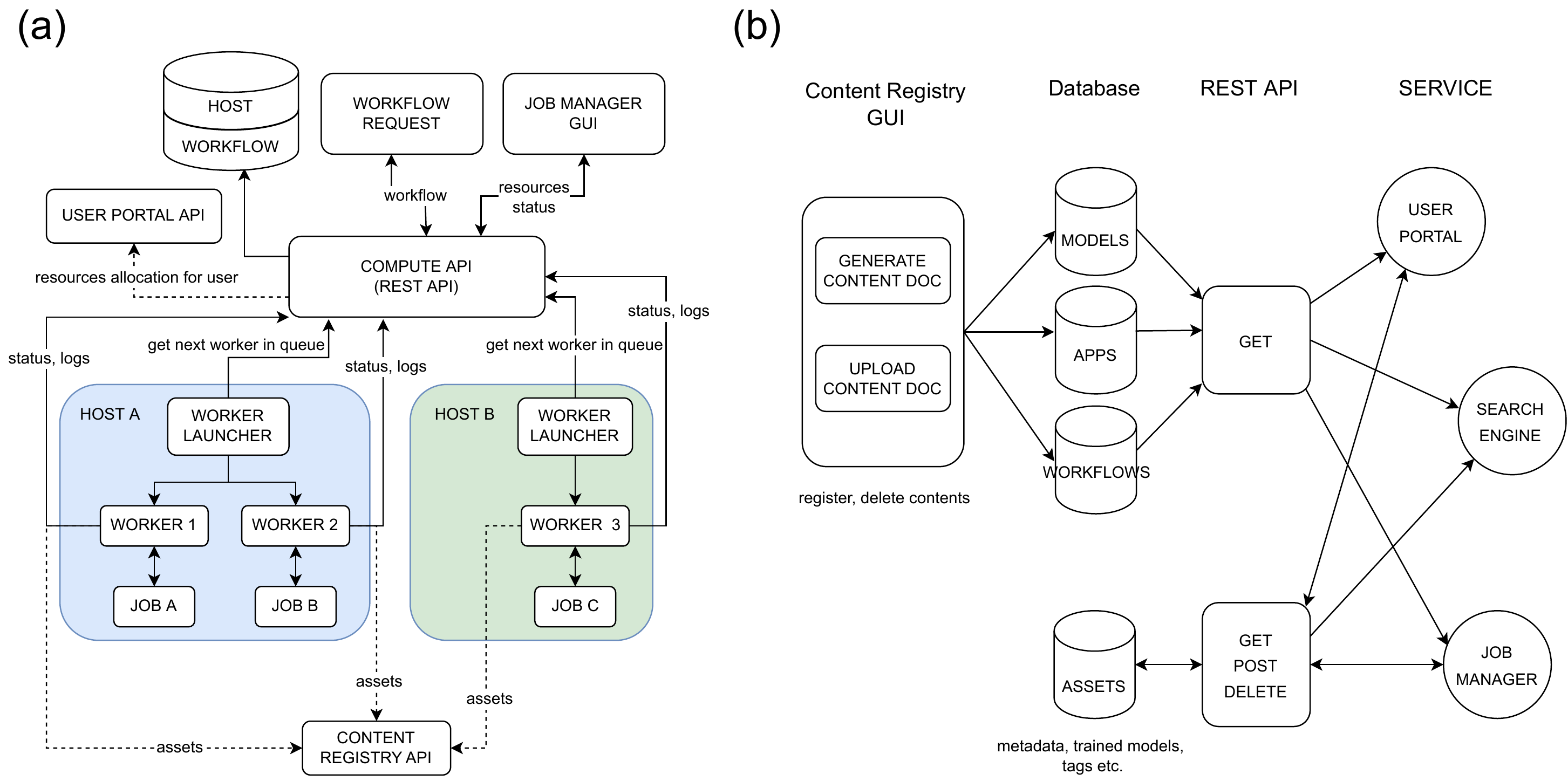}
\caption{(a) The core of the job manager, the compute API, orchestrates MLExchange workflows through a database and worker launchers. The solid lines represent the current implementation of this component, and the dashed lines correspond to new features in the next release; (b) Schematics of the content registry and its APIs. Arrow pointing right means the operation can only fetch content(s) from the upstream, whereas pointing left denotes altering content(s) in the upstream.}
\label{fig: diagram}
\end{figure*}

The scientific user facilities (SUFs) of the Department of Energy (DOE) are capable of producing over 10 petabytes of experimental and simulated data per year, making them among the biggest data producers in the world \cite{schwarz2019data, hu2021design}.
The data span multidisciplinary sciences covering multifaceted and complex interactions that require domain expertise to decipher intricate relationships within natural phenomena.
Fortunately, new advances in scientific machine learning (ML) offer an opportunity to leverage the commonalities, scientific insights, and collected experience of the larger scientific user facility community.
However, the challenge for the community to use ML in scientific studies is 3-fold:
(1) a majority of the available ML tools require users to have programmatic coding experience and a considerably profound ML background that many domain scientists do not necessarily have;
(2) most beamline data are multimodal image and spectral data that need interactive features (such as interactive visualization) in ML analyses;
(3) coordinating these complex data analyses and optimizing the use of experimental facilities and computing resources is much needed, which requires accessible machinery combining scientific ML ecosystems and user management.
Although some open-source software, such as MLflow, can achieve the concept of Machine Learning Operations (MLOps), they do not fulfill all the above needs. 
Therefore, we are building a user-friendly platform called \emph{MLExchange} to allow scientific ML algorithms and applications to be readily exchanged and deployed across the DOE facilities.
It minimizes the user barrier by providing a fully interactive experience---all services and applications are web-based interfaces that streamline user management, data pipeline, job executions, and data visualization. 
It has high deployability for each granular service, allowing it to accommodate various deployment situations.  
Furthermore, the platform has high modularity, scalability, and easy accessibility.
\medskip

\noindent\textbf{Versatile Deployment}
Every MLExchange service (a platform component, an algorithm, an frontend application, \etc) is a container \cite{merkel2014docker} that contains the required environment to run independently across different hardware.%
\footnote{MLExchange currently uses Docker containers that run on both ARM and AMD architectures. Other containers, such as Shifter containers, will be included in the next version.} 
This enables MLExchange services to be easily deployed at different locations.
For example, the official MLExchange platform is running at Vaughan (located at the Advanced Light Source, Berkeley Lab), whose resources are open to DOE facility users.%
\footnote{A NGINX proxy system is used to enable secure access to MLExchange frontend services on Vaughan.}
Moreover, users can deploy the whole platform or its individual services within their internal networks and not share their resources.
\medskip

\noindent\textbf{Modularity}
MLExchange can construct ML workflows of any complexity.
A workflow is a combination of frontend apps (\eg, a Dash app) and/or backend jobs (\eg, a ML algorithm). 
For instance, a typical workflow would consist of a few frontend apps to form an analysis, where each app could use a list of available backend algorithms.
With the MLExchange content registry, users can ingest these pre-defined workflows, new algorithms, and apps into the platform, thus populating use cases (examples are discussed in \cref{sec: use cases}).
\medskip

\noindent\textbf{Scalability}
There are two aspects of MLExchange scalability.  
First, MLExchange is designed to accommodate a large user base.
Its user portal is responsible for registering new users, authenticating user identity, and authorizing access to services and resources.
In terms of computing, the job manager is designed to handle multi-job situations, \eg, allocating computational resources for individual jobs.
Second, because of the high modularity, the user community can scale up MLExchange applications and use cases.
\medskip

\noindent\textbf{Accessibility} 
MLExchange is a web-based platform. 
Users can access its official version through \href{https://mlexchange.als.lbl.gov}{\nolinkurl{https://mlexchange.als.lbl.gov}}. 
\medskip

With the above design features, we expect MLExchange to cater to the needs of beamline scientists and facility users under various situations.
Moreover, other entities or companies may also find MLExchange a valuable platform.

\section{Architecture}
As is shown in \cref{fig: architecture}, the platform has 5 major components: the job manager, content registry, user portal, search engine, and data connector.
The job manager is expected to be deployed at different servers to leverage various computing resources (\emph{central} to each server).
While the content registry, user portal, and search engine remain \emph{centralized} even though users can also have a local copy of them if needed.

In MLExchange, an entity is treated as either a job or content.
A job is a running program using computing resources (CPUs and GPUs). 
A ``static'' content, \eg, a (learned or unlearned) ML algorithm, a Dash frontend, a computing resource, or metadata produced by a finished job, is stored and managed by the content registry.%
\footnote{User information and their hierarchical relationships are stored in the user portal graph database.}
Whereas a ``dynamic'' content associated with a running job, \eg, a set of input parameters for a running model or job logs, is stored and managed by the job manager.
Note that a static context is used as the default throughout the paper when referring to a content.
The search engine offers quick and relevant content searching services,%
\footnote{Contents, such as models and applications, could be used by authorized users and in many use cases. The search engine can give users recommendations about the most relevant MLExchange tools.}
and the data connector is responsible for securely transferring data to different locations.
The user portal authorizes access to the above services, then jobs from those services will be carried out by the job manager on behalf of the individual user. 

All these components can communicate through application programming interfaces (API) endpoints, each using a unique resource locator (URL).

\subsection{Job manager}
The job manager is a central job scheduler that coordinates job executions according to the availability of computing resources and services requested by the other MLExchange components.  
As is shown \cref{fig: diagram} (a), the job manager currently consists of an API service (compute API), a database, a worker launcher per host, and the job manager graphical user interface (GUI). 

The job manager executes workflows at the top of the hierarchy.
A workflow is constructed as a list of jobs with their corresponding dependencies, a pre-defined set of computing resources (CPU/GPU), and the number of workers.
The workflow execution involves allocating compute resources to the workers, where the compute API uses a supply-constrained definition setup to schedule workers through constraint programming and Satisfiability (cp-SAT) methods \cite{cpSAT}.

The compute API distributes workers among worker launchers (hosts) when a workflow is received.
Then, the worker launchers will launch worker containers to execute workflow jobs in the background.
The worker launchers periodically communicate with the compute API looking for compute resources to deploy the next workers in the queue. 
Each worker runs their assigned jobs---one at a time---according to their dependency setup and reports their current status, logs, and newly-generated assets back to the compute API.%
\footnote{
This is a crucial feature of MLExchange: the job manager automatically tracks experiment-dependent hyperparameters in the compute API and displays them (with descriptions) in the application interface (job table component).
Users can also permanently store this information in the content registry database.
}
Once a worker completes its job list, it will terminate itself and release the computing resources back to the host (by updating the number of available resources in the database). 

\subsection{Content registry}
The content registry consists of a GUI, a centralized database, and a set of APIs to manage the ingestion, removal, and pulling of these contents, see \cref{fig: diagram} (b).
The database currently stores four types of contents, \ie, models (algorithms), applications (apps), workflows, and assets (metadata such as service parameters).

Currently, the content of the model/app/workflow type can be registered using the GUI---by either filling out the required forms (which also generates the content document) or uploading an existing content document (JSON file)---and retrieved through API calls.
In addition, the GUI displays the available models/apps/workflows in a table where users can navigate and delete contents and submit services (execute or stop containers) of the selected contents to the job manager. 
Whereas the content of assets can be added, deleted, and retrieved to/from the content registry database only through API calls.

Note that the content registry does not store the actual content.
Instead, it stores the specifications of the content and a pointer (URL) to the place where the content is actually stored.

\subsection{User portal}
The user portal provides user management services at both the administrator and participant levels. 
From a generic perspective, it is responsible for user authentication and authorization of the MLExchange services. 
We use attribute-based access control (ABAC) \cite{ahmadi2019graph} implemented over a Neo4J graph database to enable flexible yet powerful formalization of access policies across a variety of access management cases. 
Using a graph database that we query with Cypher \cite{cypher2018SIGMOD}, we can efficiently query and analyze connected data over a set of complex policies to enforce hierarchical access to MLExchange content registry and user-owned assets.%
\footnote{Common examples of user-owned assets include but are not limited to trained ML models, workflows, and privately-owned computing resources.} 
Management of owned assets can be executed at a participant (single user), owner-defined team (group of users), or community-level (all users).

\subsection{Search engine and data connector}
Two other essential MLExchange services are content searching and ranking through the search engine, and secure data transfer via the data connector. 
The search engine uses ElasticSearch to provide fast and high-volume content searching answers. 
It also contains a reverse image searching pipeline \cite{Araujo_etal2018} that allows users to search for similar images based on deep neural network algorithms.
The data connector is currently under development. 
It will contain a GUI where users would have a collection of secure data transfer technologies to use, such as Secure File Transfer Protocol (SFTP), Secure Copy Protocol (SCP), and Globus.

\section{Use cases}
\label{sec: use cases}
We expect MLExchange to execute workflows for any scientific application. 
Two available use cases are presented in this section.

\subsection{Image segmentation}
\label{sec: seg demo}

We adapted the image segmentation workflow from Dash Enterprise App Gallery \cite{DashAppGallery, plotly2020segmentation} and successfully integrated it into the MLExchange platform.
In addition to interactively uploading new datasets, more user-friendly workflow features are added to the application.

\begin{figure}[h!]
\centering
\includegraphics[width=\linewidth]{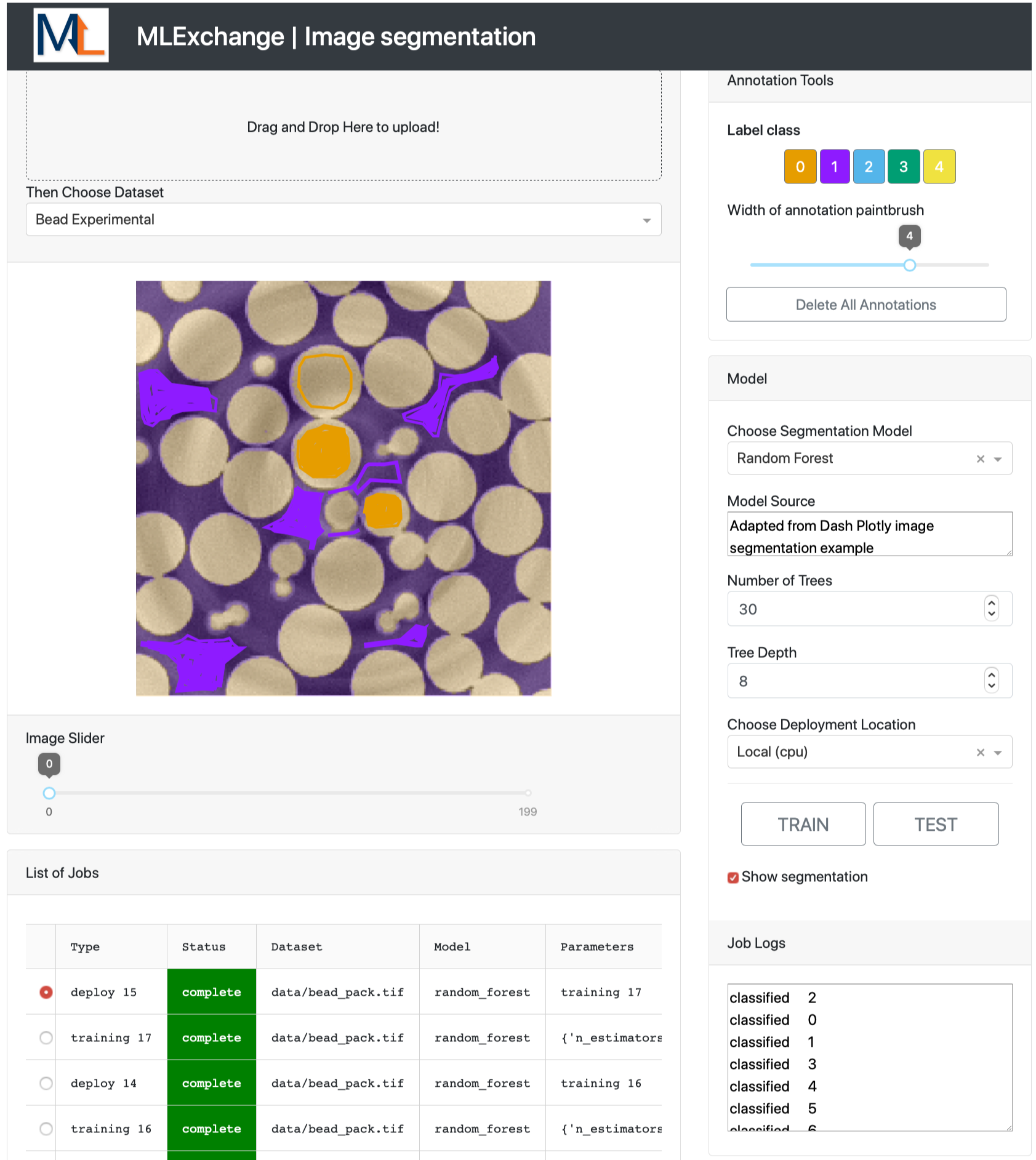}
\caption{MLExchange image segmentation user interface. It shows the segmentation results from using the Random Forest with an overlay of label shapes.}
\label{fig: seg demo}
\end{figure}

\noindent\textbf{General Procedures}
Users need to register a new ML algorithm in the content registry before using it from the user interface.
The interface supports labeling, training models, and segmenting images, see \cref{fig: seg demo}.
All models follow a TRAIN--TEST procedure: a learned model is saved in the TRAIN step and will be used to segment the full image stack in the TEST step. 
Although labeling is not needed for an unsupervised model (K-means) in the TRAIN step, users can select a portion of the images as inputs for the unsupervised learning in this step.
In the Show Segmentation mode, pixels will be colored according to the corresponding label class. 
\medskip

\noindent\textbf{ML Models}
Currently, the image segmentation application offers the following ML algorithms: decision tree-based Random Forests \cite{DashAppGallery} for supervised learning, K-means clustering for unsupervised learning \cite{scikit-learn}, and Mixed-Scale Dense Convolutional Networks (MSDNets) \cite{pelt2018improving, pelt2018mixed, goodfellow2016deep, bejani2021systematic}, a deep, fully convolutional neural network architecture that leverages dense interconnectivity between all network layers to drastically reduce the number of learnable network parameters and alleviate overfitting.%
\footnote{MSDNets are administered via pyMSDtorch (\href{https://pymsdtorch.readthedocs.io}{\nolinkurl{https://pymsdtorch.readthedocs.io}}), an open-source, Python-based deep learning library for scientific image analysis.}
\medskip

\noindent\textbf{Adaptive GUI Components}
The model parameter layouts are automatically generated and updated when selecting a different model. 
The keywords to describe these Dash components are pre-defined in the content registry.
A code takes these keywords and updates the children property of the respective Dash components such that the layouts in the Dash user interface are automatically refreshed.  
This is a key MLExchange feature as it allows users to ingest new algorithms without the need to modify the source code of their frontend applications. 
\medskip

\noindent\textbf{Workflow Execution}
In this use case, a workflow only has one job per user request (TRAIN or TEST).
\medskip

\subsection{Image labeling}
\begin{figure}[h!]
\centering
\includegraphics[width=\linewidth]{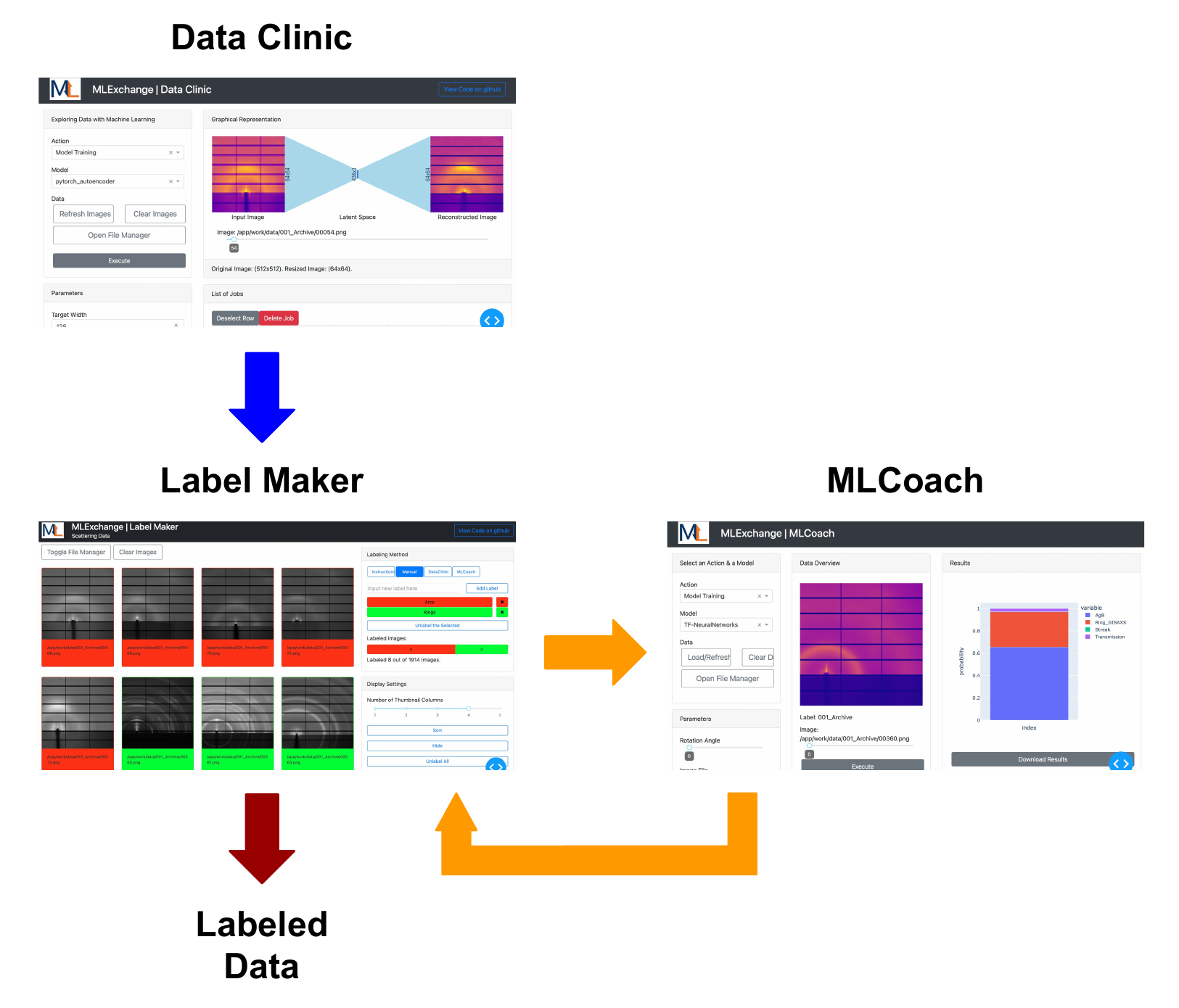}
\caption{MLExchange image labeling pipelines: unguided manual labeling through (1) Label Maker, and ML-assisted labeling through (2) Dataclinic (self-supervised) \(\to\) Label Maker; (3) Label Maker (manual labeling) \(\to\) MLCoach (supervised) \(\to\) Label Maker (auto labeling)}
\label{fig: labelmaker pipeline}
\end{figure}

In contrast to the single application case (image segmentation in \cref{sec: seg demo}), the image labeling pipelines streamline the labeling process for large datasets with three ``standalone'' frontend applications, \ie, Label Maker, Data Clinic, and MLCoach, see \cref{fig: labelmaker pipeline}.
Label Maker has an integrated interface for manual labeling and ML-assisted labeling steps using the results from the other two applications.
\medskip

\noindent\textbf{Self-supervised Learning Approach}
Data Clinic provides the similarity score (the Euclidean distance between two latent vectors) for a given target image against each image within the dataset through a well-trained autoencoder \cite{rumelhart1986learning, lecun1998gradient}.
Label Maker exploits such results by arranging the images from the most similar to the least alike, where users can proceed to label them in batches.
\medskip

\noindent\textbf{Supervised Learning Approach} MLCoach trains neural networks with a set of previously labeled images to classify the whole dataset, where the classifier's output is the prediction probability per class.
Then, Label Maker can automatically label each image on condition that its predicted probability exceeds a user-defined threshold for a specific class (\eg, \SI{50}{\percent}).
\medskip

\noindent\textbf{Workflow Execution}
The image labeling pipelines have three standalone frontend applications that can be launched in parallel or any order.
For each application, a workflow works the same way as the image segmentation application.
\medskip

\section{Summary and Outlook}
MLExchange is a web-based platform that manages the full life cycle of ML tools which is accessible and easy to use for beamline scientists. 
So far, we have built 4 major components, \ie, the job manager, content registry, user portal, and search engine.
The last major component, the data connector, is under development.
Additionally, we have implemented an assortment of web applications for scientific analysis of multimodal datasets, such as grain/pattern orientation detection, latent space exploration, peak detection for 1-dimensional X-ray diffraction (XRD) data, inpainting detector gaps in X-ray scattering \cite{inpainting}, and fast artifact identification for raw XRD images  \cite{yanxon2022artifact}.

Future development of the MLExchange platform will be focused on optimizing user experience (completing the data pipelines, supporting more container technologies, \etc) and including more use cases for scientific research.

\section*{Acknowledgment}
The authors acknowledge financial support by the U.S. Department of Energy through Collaborative Machine Learning Platform for Scientific Discovery (award No. 107514).

\section*{Code availability}
The MLExchange platform source codes can be found at \href{https://github.com/mlexchange}{\nolinkurl{https://github.com/mlexchange}}.

\bibliographystyle{IEEEtrans}
\bibliography{xloop2022}

\end{document}